\documentclass[sigconf]{acmart}
 \usepackage{color,soul}
 \usepackage{stfloats}
\AtBeginDocument{%
  \providecommand\BibTeX{{%
    \normalfont B\kern-0.5em{\scshape i\kern-0.25em b}\kern-0.8em\TeX}}}

\setcopyright{acmcopyright}
\copyrightyear{2021}
\acmYear{2021}

\acmBooktitle{2021 Symposium on Eye Tracking Research and Applications (ETRA '21 Full Papers), May 25--27, 2021, Virtual Event, Germany}
\acmPrice{15.00}

\setcopyright{acmlicensed}\acmConference[ETRA '21 Full Papers]{2021 Symposium on Eye Tracking Research and Applications}{May 25--27, 2021}{Virtual Event, Germany}
\acmPrice{15.00}
\acmDOI{10.1145/3448017.3457387}
\acmISBN{978-1-4503-8344-8/21/05}

\acmSubmissionID{1021}

\begin{document}


\title{Where Do Deep Fakes Look? Synthetic Face Detection via Gaze Tracking}

\author{\.Ilke Dem\.ir}
\orcid{0000-0003-4177-0311}
\affiliation{%
  \institution{Intel Corporation}
  \city{Santa Clara}
  \state{CA}
  \country{USA}
}
%\email{ilke.demir@intel.com}

\author{Umur Aybars \c{C}\.ift\c{c}\.i}

\affiliation{%
  \institution{Binghamton University}
  \city{Binghamton}
  \state{NY}
  \country{USA}
}
%\email{uciftci@binghamton.edu}
\renewcommand{\shortauthors}{Demir and \c{C}ift\c{c}i}


\begin{abstract}
Following the recent initiatives for the democratization of AI, deep fake generators have become increasingly popular and accessible, causing dystopian scenarios towards social erosion of trust. A particular domain, such as biological signals, attracted attention towards detection methods that are capable of exploiting authenticity signatures in real videos that are not yet faked by generative approaches. In this paper, we first propose several prominent eye and gaze features that deep fakes exhibit differently. Second, we compile those features into signatures and analyze and compare those of real and fake videos, formulating geometric, visual, metric, temporal, and spectral variations. Third, we generalize this formulation to the deep fake detection problem by a deep neural network, to classify any video in the wild as fake or real. We evaluate our approach on several deep fake datasets, achieving 92.48\% accuracy on FaceForensics++, 80.0\% on Deep Fakes (in the wild), 88.35\% on CelebDF, and 99.27\% on DeeperForensics datasets. Our approach outperforms most deep and biological fake detectors with complex network architectures without the proposed gaze signatures. We conduct ablation studies involving different features, architectures, sequence durations, and post-processing artifacts.
\end{abstract}





  \begin{teaserfigure}
    \includegraphics[width=1\textwidth]{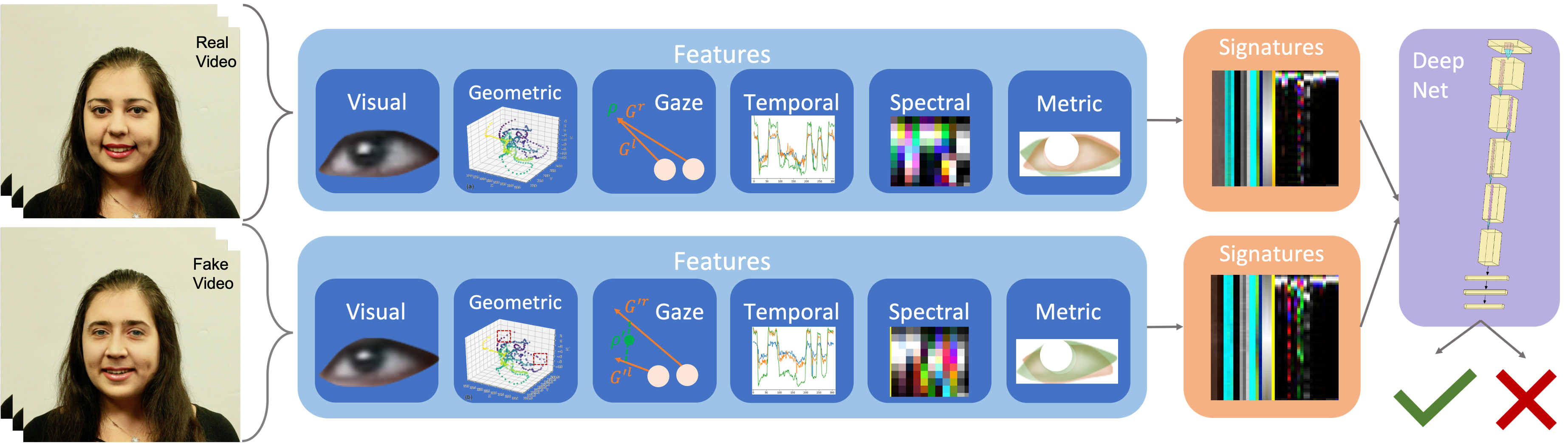}
    \caption{\textbf{Overview.} Our fake detector extracts eye and gaze features from real (top) and fake (bottom) videos. Frame-based features (blue) are converted to temporal signals and spectral densities to forge signatures (orange). Finally, a deep network (purple) predicts authenticity from signatures, and aggregates predictions.}
    \Description{Gaze-based deep fake detector overview.}
    \label{fig:teaser}
  \end{teaserfigure}

\maketitle

\section{Introduction}
With the growth of computational resources and the availability of deep learning models, AI applications thrive on finding new use cases. In particular, generative models invigorate by creating more photorealistic results with each technological advancement. One particular example is in face synthesis domain: The resolution and potent of generated faces have been increasing exponentially since the introduction of Generative Adversarial Networks (GANs)~\cite{gan}. Although realistic synthetic faces are promoted for mixed reality applications, avatars, digital humans, and games; detrimental uses also emerged, such as deep fakes.

Deep fakes are photorealistic synthetic portrait videos generated by deep neural networks, in which the main actor is replaced or reanimated with another person or with a synthetic face. Deep fakes emerged as a novel technique in facial reenactment~\cite{FaceSwap} and rapidly became a mainstream approach for media manipulation. Consequences are already faced or forecasted, such as impersonation~\cite{cnnblog}, celebrity porn~\cite{deepfakereddit}, and political misinformation~\cite{threatblog}. Moreover, interdisciplinary groups and governments provided action plans~\cite{ucla} and acts~\cite{bill} as a precaution for deep fakes.

To combat malicious use of deep fakes, several detection approaches are proposed, mainly categorized as pure deep learning methods and semantic methods. Deep methods classify authenticity only based on the input image/video/audio dataset, without an intermediate representation. Semantic approaches exploit some priors hidden in real or fake data, such as blinks~\cite{blink}, heart rates~\cite{fakecatcher}, phoneme-viseme mismatches~\cite{phovis}, or eye reflections~\cite{cornealspec}. First, we observe that the eye-related fake detectors only consider single artifacts of real eyes, as blinks or reflections. Second, by inspecting digital humans in production, we conclude that inconsistent gaze directions, vergence points, and eye symmetries create the "uncanny valley" for synthetic faces. We want to analyze and integrate these variants in a deep fake detector.

In this paper, we first examine the spatial, temporal, and spectral consistency of eyes and gazes in five domains: (i) The visual domain to cover color and shape deformations, (ii) the geometric domain to include characteristics of 3D vergence points and gaze directions, (iii) the temporal domain to assess the consistency of all signals, (iv) the spectral domain to investigate signal and noise correlation, and (v) the metric domain to learn from the deviation of the coupled signals as approximated by the generators. The rest of the paper discloses our classification network that collates described features into a deep fake detector. Our main contributions are listed below.
\begin{itemize}
 \item We present \textit{the first} holistic analysis on deep fake eyes and gazes; proposing geometric, visual, temporal, spectral, and metric features.
 \item Based on our analysis, we exploit inconsistencies of synthetic eyes to build a robust deep fake detector.
 \item We achieve significant accuracies in our experiments on different fake generators, with different visual artifacts, and on unseen deep fakes in the wild; using \textit{only} the proposed eye and gaze features.
\end{itemize}
We evaluate our approach on four publicly available datasets, namely FaceForensics++~\cite{FF++}, CelebDF~\cite{Celeb_DF_cvpr20}, DeeperForensics~\cite{deeperforensics}, and Deep Fakes Dataset~\cite{fakecatcher}, achieving 92.48\%, 88.35\%, 99.27\% and 80.0\% detection accuracies respectively. We conduct ablation studies with different generators, network architectures, sequence lengths, and feature sets. We compare our approach against three biological detectors and six learning-based detectors on three datasets. Finally, we perform experiments including (1) cross validation between dataset splits, (2) cross-dataset evaluations,  (3) weighing the contribution of different feature domains, and (4) assessing the robustness against post-generation image manipulation.

\section{Related Work}
We briefly introduce face synthesis, then delve into detection methods. Although forgery detection is an established domain with immense literature, we limit our compilation to deep fake detection, especially with biological priors.

\subsection{Parametric Face Synthesis}
The proliferation of deep generative models enabled realistic face creation and manipulation techniques such as reenactment~\cite{f2f, deepvideoportrait, vid2vid}, attribute or domain manipulation~\cite{stargan20, attgan17, dfn}, expression molding~\cite{exprgan}, and inpainting~\cite{gfc}. In particular, physically grounded models~\cite{Saito2017PhotorealisticFT, hao20, flame} generate realistic humans by learning facial parameters. Although there is a few techniques focusing on gaze reenactment~\cite{ganin16,Thies:2018:FRG:3191713.3182644}, artifacts remain when head is not front-facing and inconsistencies exist between pose and perspective. All of these approaches still lack consistent gazes as (1) their approximations are not accurate at the gaze level, (2) the gaze signatures are not well-replicated in spatial (i.e., misalignment), temporal (i.e., skip frames), and spectral (i.e., noise) domains, and (3) detailed eye models are not encoded in their parametrizations. We aim to use gaze consistency as an authenticity indicator of real videos.

\subsection{Fake Detection with Biological Priors}
Deep fake detectors either employ complex networks to classify videos without an intermediate representation, or they make use of simpler networks with some semantic interpretation of authenticity signatures or generative residuals. The first family of detectors have limitations such as only processing images~\cite{shallownet,8014963,8124497,8682602} or audio~\cite{8553270}, detecting the fakes of specific generators~\cite{mesonet, shallownet, 8553251,dctforgery} or specific persons~\cite{Agarwal_2019_CVPR_Workshops}, or using small datasets~\cite{8638330,Li_2019_CVPR_Workshops}. We refer the reader to a recent survey~\cite{all} for the details of each network. The second branch extracts intermediary physiological, biological, or physical signals, then learns the authenticity from their correlations. Such approaches enclose blinks~\cite{blink,blink2}, head pose~\cite{8683164}, eye reflections~\cite{cornealspec}, heart rates~\cite{fakecatcher, ijcb, cgppg}, facial attributes~\cite{8638330}, depth of field~\cite{dofnet}, eyebrows~\cite{eyebrow}, phoneme-viseme mismatches~\cite{phovis}, and affective cues~\cite{affective}. Most of these approaches need some portion of the skin to be visible, decaying in accuracy under heavy make-up or insufficient illumination. Being invariant to skin reflectance, we aim for a more robust approach under these conditions to outperform and/or complement other fake detectors.

\begin{figure*}[h]
\centering
  \includegraphics[width=1\linewidth]{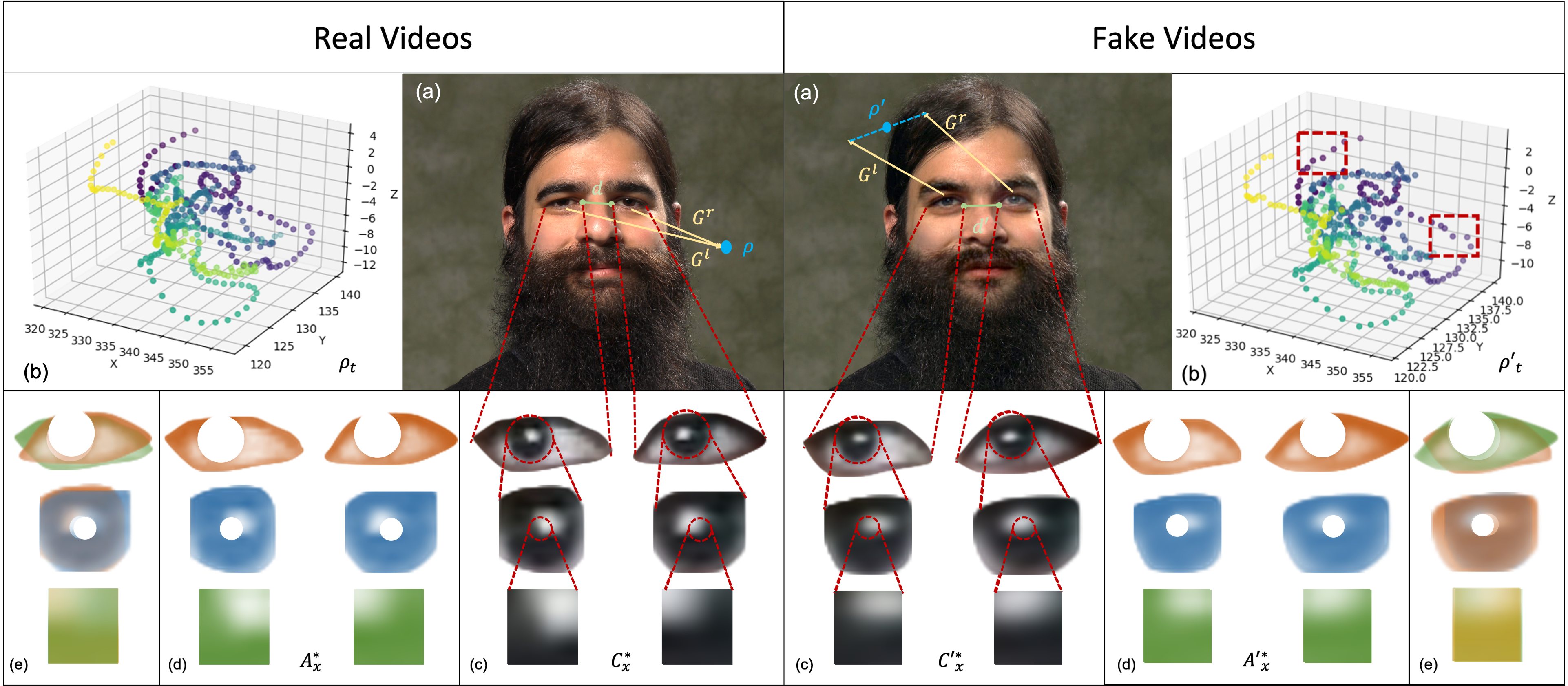} 
 \caption{ \label{fig:features}Eye and Gaze Features. Real (left) and fake (right) videos (a), exhibit different geometric (a,b), visual (c,d), and metric (e) features. We draw gaze vectors $G^*$, 3D gaze points $\rho$, and pupil distances $d$ on (a).} 
\end{figure*}

\subsection{Gaze-based Fake Detection}
Recently,~\cite{blink} proposes a robust detector by monitoring irregular blinking patterns successfully. Further investigation of eye traits reveals that the corneal specular highlights for real eyes have strong similarities while those for GAN-synthesized eyes are different.~\cite{cornealspec} introduces a physiological/physical detection method based on that observation. However, this approach exposes only one feature of eyes and it is still sensitive to both illumination and pose variations. 

In addition to blinking and corneal reflection patterns, gaze is also an important trait to distinguish real and fake videos. Nevertheless, there is no existing work to utilize the eye gaze discrepancy for identifying deep fakes. This paper is to assemble such important but subtle signals to tackle the deep fake detection problem. 

To address this issue, essential steps include eye landmark extraction and gaze estimation. In general, eye detection techniques cover (1) shape-based~\cite{Hansen05}, (2) appearance-based~\cite{Hansen10}, or (3) hybrid detection~\cite{Hansen02,santini18}. Gaze estimation methods rely on 2D features~\cite{Hansen10} or 3D eye models~\cite{Reale13, Hansen10, dierkes19}. Recent developments using CNN based gaze estimation has also improved gaze estimation accuracy~\cite{openfacegaze,cnngaze,cvprgaze,mpiigaze}. In this paper, we utilize~\cite{openface} for extracting eye landmarks, and~\cite{openfacegaze} for extracting pupil and iris landmarks in addition to estimating the gaze direction. We also experiment with other approaches and implementations, however they do not provide iris tracking~\cite{mpiigaze}, require special software/hardware~\cite{opengazer, pupillabs}, calibration~\cite{pygaze}, or configured for mobile~\cite{cvprgaze} or online~\cite{gazepointer} users.

\section{3D Eye and Gaze Features}
\label{sec:features}
In order to learn the characteristics of fake gazes, we investigate several eye and gaze features and their contributions to differentiate authenticity. We explore these signals in five subsets, although some features belong to multiple domains. In addition to Figure~\ref{fig:teaser} briefly depicting our proposed features, Figure~\ref{fig:features} provides examples for all of them. We also categorically document all features in Table~\ref{tab:sign} including their normalizations and contributions to the signatures.

\subsection{Visual Features}
Visual features $V$ include color $C_*$ and area $A_*$ of eye $X_E$, iris $X_I$, and pupil $X_P$, from left $X_*^l$ and right $X_*^r$ eyes:
\begin{equation}
    V=\{C_I^r, C_I^l, C_P^r, C_P^l, A_I^r, A_I^l, A_P^r, A_P^l, A_E^r, A_E^l, |C_I^l-C_I^r|, |C_P^l-C_P^l|\}\label{eqn:v}
\end{equation}
As shown in Figure~\ref{fig:features}d, areas correspond to the pixels that are included in the larger region (i.e., iris) but not in the smaller region (i.e., pupil). The color is computed in the CIElab~\cite{cielab} space in order to create illumination independent features, then averaged within the corresponding area (Figure~\ref{fig:features}c).
Assuming that the color difference of iris and pupils from left and right eyes would stay constant (in an illumination invariant color space), we also append color differences (Eqn.~\ref{eqn:v}). Note that we excluded the eye color as it is brittle against illumination and alignment errors. 
\begin{figure*}[ht]
\centering
\includegraphics[width=1\linewidth]{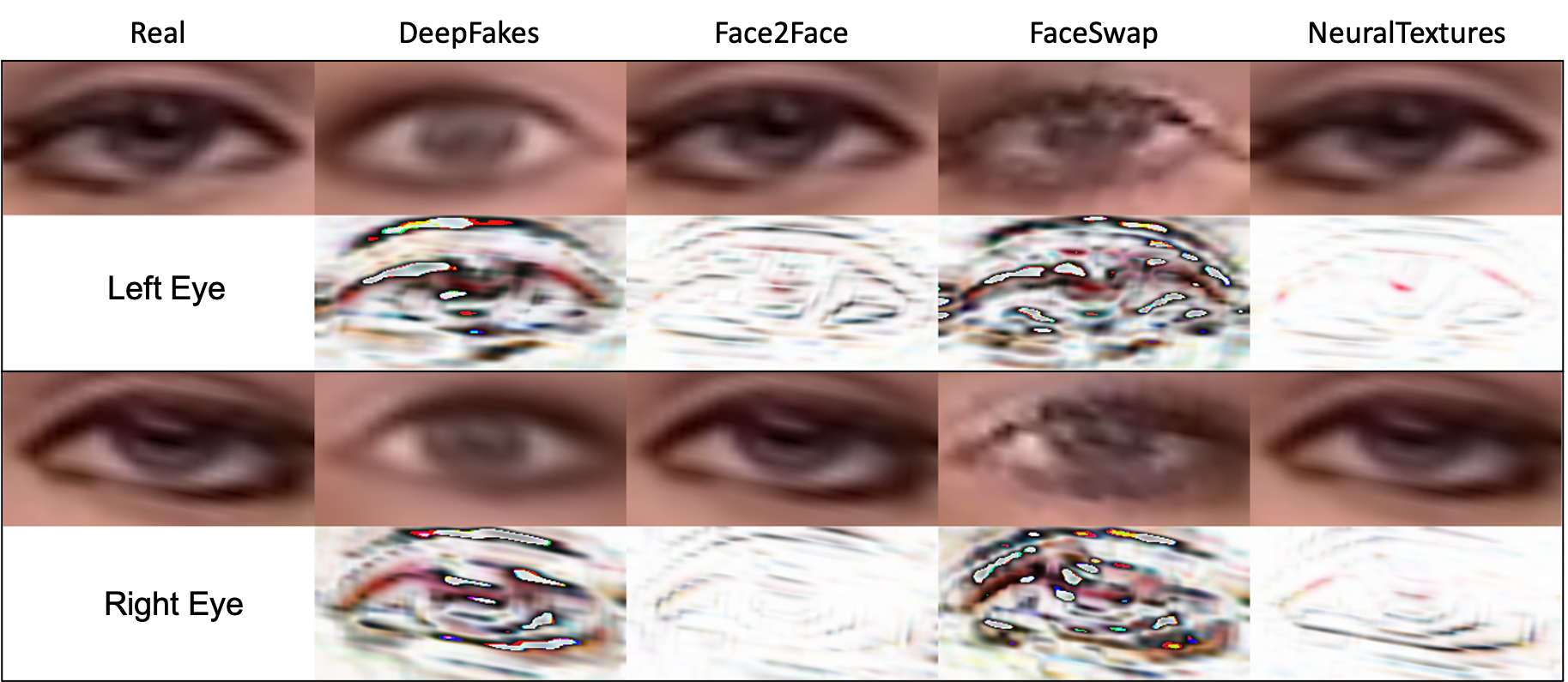}
\caption{\textbf{Eye Deformations.} Real (first column) and fake (others) eyes are demonstrated to emphasize color differences (even rows) and size differences (i.e., the areas of eye regions in the second column).}\label{fig:visual}
\end{figure*}

\subsection{Geometric Features}
Geometric gaze features $L_G$ include left and right gaze vectors $G^l, G^r$, 3D gaze point $\rho$, 3D gaze point approximation $\hat{\rho}$, and the value of the cost function at the approximation $\delta_{\rho}$. Geometric eye features include the eye distance $d$, pupil distance $d_p$, and the difference of areas $|A_*^l-A_*^r|$:
\begin{align}
    L_G&=\{G^l, G^r, \rho,\hat{\rho}, \delta_{\rho}\}\\
    L_E&=\{d, d_p, |A_E^l-A_E^r|, |A_I^l-A_I^r|, |A_P^l-A_P^l|\}
\end{align}
We use~\cite{openfacegaze} to extract gaze vectors and pupil centers. Then we shoot rays from the pupil centers along the gaze vectors to find their intersection. We aim to optimize for the mid-point of the closes points of the two rays, because it is possible that the rays are skew lines in 3D. In order to approximate the intersection of two 2D rays in 3D space, we use a least squares optimization with $1e-08$ threshold. In addition to the approximated 3D gaze point $\rho$ (green points in Figure~\ref{fig:features}a), we compute the distance between the projections of $\rho$ onto $G^l$ and $G_r$ (orange vectors in Figure~\ref{fig:features}a), indicated as $\hat{\rho}$ (the green line in fake Figure~\ref{fig:features}a). We also incorporate the cost at the solution as $\delta_\rho$.

\subsection{Temporal Features}\label{sec:temp}
As of now, the features we present are computed per frame. However we are less interested in their exact values but more interested in their consistency, therefore we introduce the concept of \textit{sequences}: a fixed $\omega$-length of consecutive frames from a video. We divide all videos into $\omega$ length sequences and create signals from the previously defined visual and geometric features, in order to perform our temporal consistency analysis.
\begin{equation}
T = \{V^i, L_G^i, L_E^i\}\ such\ that\ i \in [0,\omega)
\label{eqn:temp}
\end{equation}
Although $\omega$ can be set as a user parameter for detector sensitivity, we experiment with different $\omega$ values in Section~\ref{sec:Experiments}.

\subsection{Metric Features}
In addition to temporal consistency, spatial coherence is an embedded property of authentic videos, especially for symmetries. We employ cross correlation $\phi(*,*)$ of left/right, eye/gaze features to utilize their spatial coherence as an authenticity cue (Figure~\ref{fig:features}e).
\begin{align}    \label{eqn:phi}
    M = \{&\phi(C_I^l,C_I^r),\phi(C_P^l,C_P^r), \\\nonumber
    &\phi(A_I^l,A_I^r), \phi(A_P^l,A_P^r), \phi(A_E^l,A_E^r),\\\nonumber  
    &\phi(G^l,G^r)\}
\end{align}

\subsection{Spectral Features}
Although we launch our analysis in temporal domain, the environment or the face detection algorithm can introduce noise which deteriorates our temporal features. Thus, we calculate the power spectral density $\Theta(*)$ of all the previously defined signals. Frequency bands play a crucial role in differentiating noise from actual gaze or eye movements.
\begin{equation}
    S = \{\Theta(V^i), \Theta(L_G^i), \Theta(L_E^i), \Theta(M) \}
\label{eqn:spec}
\end{equation}

In total, we conclude with $40x3x\omega$ size signatures with the combination of $S$ and $T$, all of which are listed in Table~\ref{tab:sign}.

\section{Gaze Authenticity}
\label{sec:anal}
Having defined our feature space, we explore and analyze the behavior of the predefined features on real and fake gazes. As different generative models create different levels of realism for synthetic faces, they also leave different traces behind~\cite{ijcb}. We experiment with several generator outputs~\cite{DeepFakes,FaceSwap-Gan,f2f,neuraltex} corresponding to a real sample. For the domains in which the artifacts of different generators are not significantly varying, we discuss them collectively.

\begin{figure*}[h]
\centering
\includegraphics[width=1\linewidth]{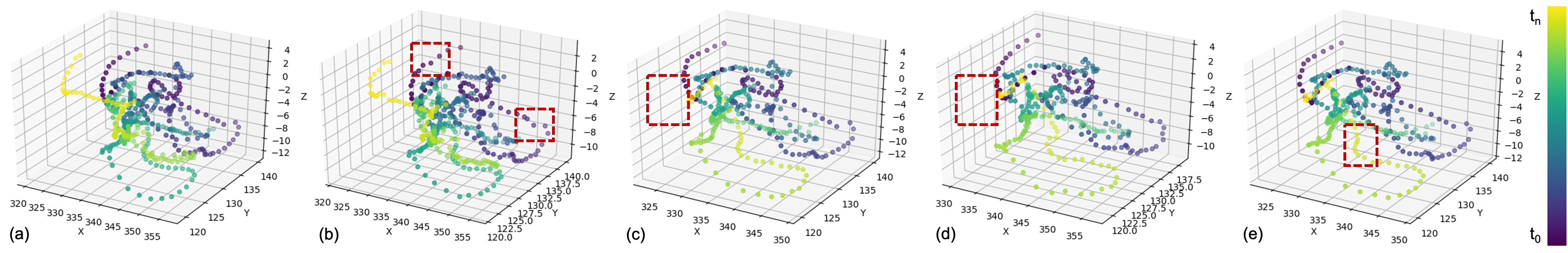}
\caption{\textbf{3D Gaze Points} of a real video (a) and its fake counterparts (b-e) are observed for $\omega$ frames of a video. Fake gazes exhibit noise (b), miss some saccades (c,d), and have irregular distribution (e). Sample problematic points are boxed in red.} \label{fig:3D}
\end{figure*}
\begin{figure*}[bp]
\centering
  \includegraphics[width=1\linewidth]{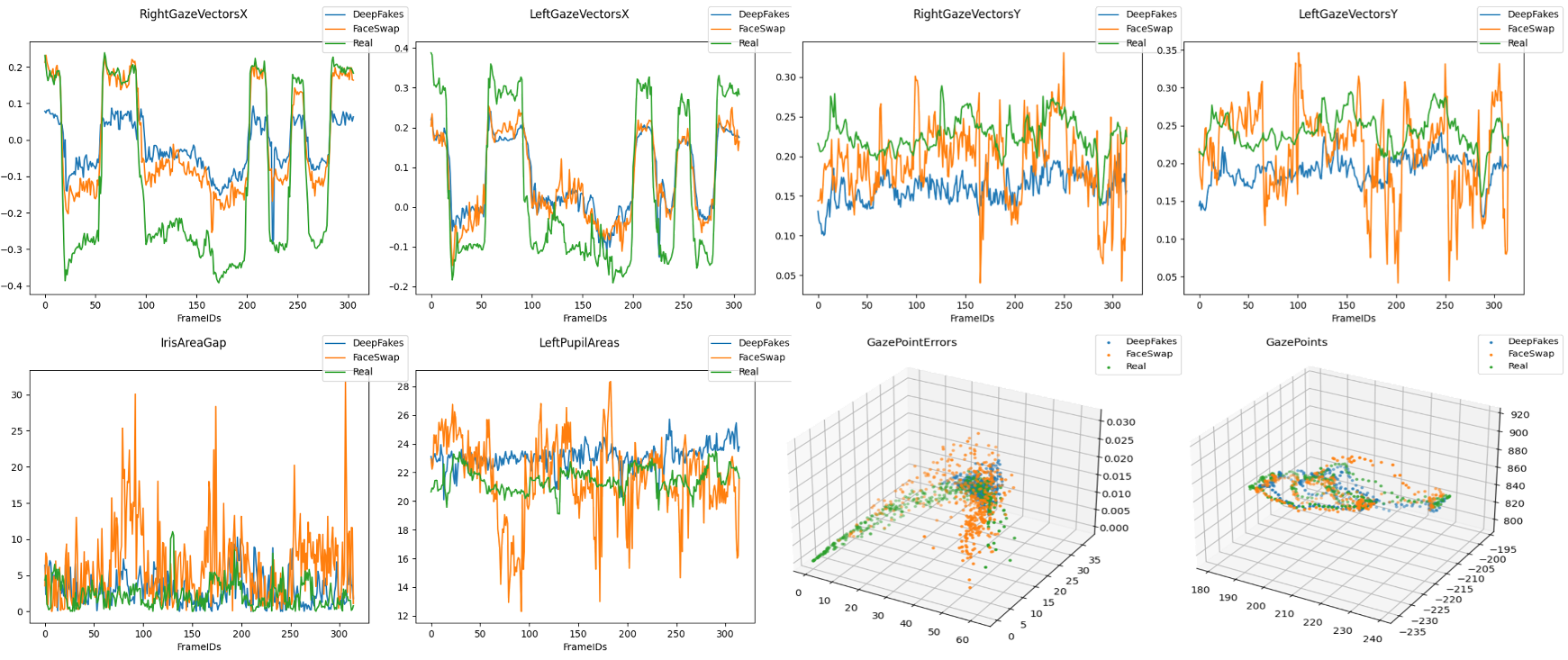}
 \caption{\textbf{Temporal Signals.} Real (green) and fake (orange and blue) signals of (a) $G^r_t[0]$, (b) $G^l_t[0]$, (c) $G^r_t[1]$, (d) $G^l_t[1]$, (e) $|A_I^l-A_I^r|_t$, (f) ${A_P^l}_t$, (g) $\hat{\rho}_t$ and (h) $\rho_t$. Fake gazes cover less range in (a, b) and they exhibit more noise in (c,d). Fake iris area differences has a $30 mm^2$ range in (e) and fake pupil areas can differ by $14 mm^2$ in (f).} 
  \label{fig:signals}
\end{figure*}

\subsection{Color and Size Deformations}
The visual features $V$, namely color and size of the eye regions, tend to stay coherent for real videos, even if there is some movement involved. Figure~\ref{fig:visual} demonstrates close ups of real (first column) and realistic fake (others) eyes of a sample from FF++ dataset. Even for the best fake generator (last column), some of the resolution, striding, or alignment artifacts become visible in close ups. We also emphasize the color differences of those real-fake pairs (even rows) and area changes of regions (i.e., the second column). We delve into those deformations to detect fake videos.

\subsection{Geometric Attestation}
A visually invisible but computationally powerful way to verify natural gazes is to geometrically validate them. Intuitively, the gaze vectors in real videos converge at a point in 3D space. However, slight deformations of pupil centers or gaze directions may lead to two completely off-plane gaze vectors that never intersect (i.e., Figure~\ref{fig:features}a fake, or Figure~\ref{fig:teaser} fake). We use vergence points to verify authentic gazes, such that (i) $\rho$ exists and is in front of the face, (ii) $\hat{\rho}=0$ or is relatively small, and (iii) $\delta_\rho \times \rho$ is close to $\rho$ within the optimization $\epsilon$. Another nice property of $\rho_t$ is that it corresponds to where we naturally look, which has its own form and properties (i.e., smooth pursuits, saccades, etc.) as shown in Figure~\ref{fig:3D}.

\begin{figure*}[h]
    \centering
    \begin{minipage}{0.48\textwidth}
        \centering
        \includegraphics[width=1\textwidth]{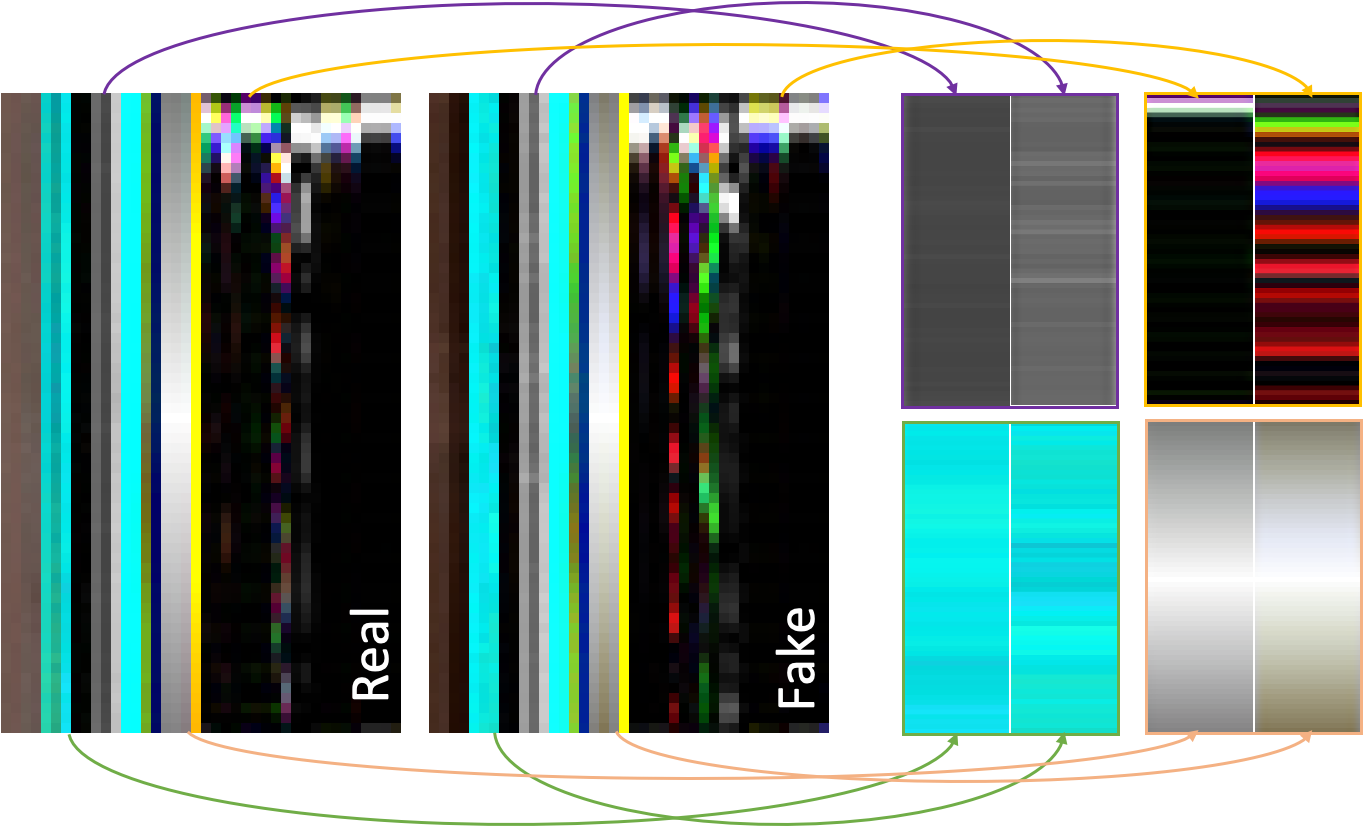} 
    \end{minipage}\hfill
    \begin{minipage}{0.48\textwidth}
        \centering
\caption{\textbf{Signatures} from real (left block) and fake (mid block) sequences are demonstrated. We zoom into four example signals: (1) Distance of eyes $d$ (first quarter, gray) almost exhibit no change for the real sequence (left), but varies for the fake one (right). (2) Eye areas $A_E^*$ (second quarter, blue) behave similar to (1). (3) PSD of real pupil areas $\Theta({A_P^*}_i)$ is significantly smaller than the fake $\Theta'({A_P^*}_i)$ (third quarter, black/colorful). (4) Cross correlation of left and right gaze vectors $\phi(G^l,G^r)$ (fourth quarter) are more uniform for reals than for fakes, as we expect a gray level gradient for the perfect match.} \label{fig:spec}
    \end{minipage}
\end{figure*}

\subsection{Temporal Consistency}
To illustrate temporal consistency of gaze features, we compare the form of 3D gaze points from real (first) and fake (others) versions of a video in Figure~\ref{fig:3D}, color coded by time. In red boxes, we observe that fake gazes exhibit noise (Figure~\ref{fig:3D}b), miss sudden motions (\ref{fig:3D}c,d), and have ragged distributions (\ref{fig:3D}e) compared to real gazes.

Similarly, we show exemplar temporal signals $G^l_t$, $G^r_t$, $|A_I^l-A_I^r|_t$, ${A_P^l}_t$, $\hat{\rho}_t$, and $\rho_t$ in Figure~\ref{fig:signals}, of real (green) and fake (blue and orange) videos. Fake gazes may break temporal consistency because of over smoothing (i.e., $G^l_t[0], G^r_t[0]$), or noise (i.e., left-right iris size changes by $30 mm^2$), or random distributions (i.e., gaze error).

\subsection{Spectral Artifacts}
Depending on video resolution, post-processing operations, and illumination; real videos may produce imperfect signals. Those imperfections differ in nature from the generative noise, and the distinctions are best observed in spectral domain. Therefore, we exert power spectral densities of our temporal signals in our analysis, similar to the detection approaches in biological domain~\cite{fakecatcher}. Synthetic features produce a richer spectra than reals, as we expect most of our real features to remain constant or coherent. Figure~\ref{fig:spec} shows the signature blocks, as well as zoom ins to several features, including PSDs of real and fake pupil areas $\Theta({A_P^*}_i)$ at top right. The colorful bands indicate different spectral densities, so the fake PSD gleams whereas the real PSD contains only a few bands.

\subsection{Symmetry Flaws}
Early signs of detecting GAN-generated images were claimed to be asymmetric eyes, ears, and earrings. As the convergence of GANs needs more resources than training other deep networks, results from sooner epochs still exhibit asymmetric eyes as there is no explicit control for coupling eyes. We exploit this shortcoming and experiment with the cross correlation of symmetric pairs of eye and gaze features, with the hypothesis of fake features breaking the symmetry. Our experiments yield that $\Phi = \{\phi(C_I^l,C_I^r), \ldots , \phi(G^l,G^r)\}$ (Eqn.~\ref{eqn:phi}) increases the detection accuracy.

\section{Learning Authenticity by Eye \& Gaze Features}
In this section, we build a deep fake detector using our features from Section~\ref{sec:features} and our analysis from Section~\ref{sec:anal}. 
\subsection{Datasets}
\label{sec:datasets}
As our hypothesis of utilizing eye and gaze features for deep fake detection needs to be tested on several deep fake generators, so we need datasets with multiple generators and in the wild samples. For this purpose, we use FaceForensics++ (FF++)~\cite{FF++} dataset with known and Deep Fakes dataset (DF)~\cite{fakecatcher} with unknown sources. FF++ dataset includes 1,000 real videos and 4,000 fake videos, generated by four generative models which are FaceSwap~\cite{FaceSwap}, Face2Face~\cite{f2f}, Deepfakes~\cite{DeepFakes}, and Neural Textures~\cite{neuraltex}. DF dataset, on the other hand, contains 140 in-the-wild deep fakes collected from online sources, thus they are generated by various unknown generators, undergone various post-processing and compression operations. Such deep fakes do not meet any quality standards and their source generative model is unknown, in addition to experiencing occlusion, illumination, etc. artifacts. 

We complement our experiments with CelebDF~\cite{Celeb_DF_cvpr20} (CDF) and DeeperForensics~\cite{deeperforensics} (DFor) datasets to further justify our claim of trusting eye and gaze features as a deep fake litmus. CDF includes 590 real and 5639 fake videos created by their own generative model. DFor contains 50000 real videos of 100 people captured under controlled conditions, and 1000 fake videos, extrapolated to 10000 videos by 10 image distortion algorithms.

\subsection{Network Architecture}

We treat authenticity detection as a binary classification problem. Our network is made up of three dense layers with 256, 128, and 64 nodes, followed by a last dense layer with 2 nodes. As discussed in the previous section, our features are representative and interpretable, nonetheless our dense architecture may tend to overfit without regularization. Therefore we include batch normalization and dropout layers in between the dense layers to prohibit the network from overfitting. We also add leaky ReLU activations to add non-linearity, and end with a sigmoid activation to obtain the outputs. The detailed model architecture can be observed in Table~\ref{tab:layers}.

\subsection{Training Setup}\label{sec:training}
After we extract feature sets $S$ (Eqn.~\ref{eqn:spec}) and $T$ (Eqn~\ref{eqn:temp}), we normalize all features to be in $[0,1)]$, by shifting and scaling with different values (Table~\ref{tab:sign}). Then we concatenate the processed signals to create our input tensors, or \textit{signatures} (Figure~\ref{fig:teaser}, orange boxes), of size $(40, \omega, 3)$. The decomposition of our input tensors are depicted on the left side of Figure~\ref{fig:spec} and in Table~\ref{tab:sign}. We use the following hyper-parameters determined empirically: Adam optimizer with a learning rate of $1e-04$, 0.3 dropout probability, batch size of 32, and 0.2 threshold for leaky ReLU layers. We train the models for 100 epochs and validate every 10 epochs.  

For training our networks, we gather equal number of real and fake videos from the aforementioned datasets, and create train and test subsets with a 70\%-vs-30\% split (unless otherwise is noted in the results section). This split is completely random for all datasets and it does not affect the results as demonstrated in Section~\ref{sec:fold}. Only for CDF, authors provide a predetermined test set of 178 real and 340 fake videos for benchmarking results, thus we use the provided subsets. For DFor, we use 1000 fake videos with no post-processing and 616 real videos with uniform lighting, frontal view, and all expressions. Each train/test pair belongs to one dataset, except cross dataset evaluation in~\ref{sec:fold}.

\subsection{Video Predictions}
As our signatures are per sequence (Section~\ref{sec:temp}), we need to aggregate sequence predictions from our network into video predictions. Although deep fakes mostly focus on stable faces, they can contain a moving actor, illumination changes, occlusions, and frames without faces. To accommodate for those irregularities, we can trust the average of the sequence predictions, e.g., if the mean of sequence predictions is closer to fake, we accept the whole video as fake. Another approach is majority voting, where each sequence of a video contributes to a fake or real vote. Instead of counting each sequence as fake or real, we can use the confidence of sequence predictions as a voting score. Or more suitably, we can level with the negative likelihood of sequence predictions contributing to the video prediction. In order to accomplish that, we use the average of log of odds of prediction confidences as the voting scheme:
\begin{equation}
\frac{1}{N_p} \sum_{i<N_p}{log(\frac{p_i}{1-p_i})}
\end{equation}

\section{Results}
We use OpenFace~\cite{openface} for eye and face tracking, OpenCV for image processing, SciPy and NumPy for scientific computing, Keras for deep learning, and Matplotlib for visualizations. We train our system on a single NVIDIA GTX 1060 GPU, and a full training loop of 100 epochs takes less than an hour. We continue with our evaluations on different datasets, from different generators, comparisons, validations, experiments, and ablation studies.

\subsection{Evaluations and Comparisons}
We evaluate our approach on several single and multi-source datasets discussed in Section~\ref{sec:datasets} and list our sequence and video accuracy percentages (abbreviated as S. Acc. and V. Acc.) in Table~\ref{tab:ClassACC}. CDF and DFor represent single generator datasets, FF++ is a known multi-generator dataset, and DF is a completely in-the-wild dataset. Our approach detects fakes with 88.35\%, 99.27\%, 92.48\%, and 80\% accuracies, respectively. We also evaluate our detector on four known generators of FF++ separately. We observe that for GANs that generate masks without eyes, such as~\cite{neuraltex,f2f}, our accuracy is low as expected -- they simply do not modify the eyes. 

\begin{table}[hb]
    \centering
        \centering
\caption{\textbf{Classification accuracies} on FF++ generators and single / multi / unknown source datasets. }
  \label{tab:ClassACC}
  \begin{tabular}{ccc} 
    \toprule
    Source & S. Acc. & V. Acc.\\
    \midrule
    Real & 80.36&91.30\\
    DeepFakes & 81.02& 93.28\\
    Face2Face & 58.18 & 59.69\\
    FaceSwap & 80.63 & 91.62\\
    Neural Tex. & 57.46& 57.02\\
    \midrule
    CDF~\cite{Celeb_DF_cvpr20} & 85.76 & 88.35 \\
    DFor~\cite{deeperforensics} & 95.57 & 99.27\\
    FF++~\cite{FF++} & 83.12 & 92.48 \\
    DF~\cite{fakecatcher} & 79.84 & 80\\
  
  \bottomrule
\end{tabular}
\end{table}

\subsubsection{Comparison with Biological Detectors}
In Table~\ref{tab:compbio}, we compare our approach against three other biological deep fake detectors built on blinks~\cite{blink}, which uses only the eye region similar to our approach; head pose, which uses similar information from~\cite{openface}; and heart rates~\cite{fakecatcher}, which uses the information from exposed skin. For a fair comparison, we use the same train and test subsets from FF++ and DF for all approaches as discussed in Section~\ref{sec:training}, where FF/FS and FF/DF corresponds to FaceSwap and DeepFakes subsets of FF++, respectively. Observing first two rows, blink patterns and head poses are no longer a strong indicator of fake content (50-67\%), as generators learn to fake these features. In contrast, heart rates (third row) contain significant authenticity clues similar to our approach (on the average 2.8\% better on FF++).

\begin{table}[h]
      \centering
   \caption{\textbf{Comparison} with state-of-the-art biological (1-3) and deep (4-5) detectors on FF and DF. Same data and procedure are used, except results~(*) from~\cite{FF++}.}
   \label{tab:compbio}
   \begin{tabular}{c|ccc}
     \toprule
     Approach & FF/DF & FF/FS & DF \\
     \midrule
     Blink~\cite{blink} & 67.14 & 54.15 & 57.69\\
     Head pose& 55.69 & 50.08 &  67.85\\

    PPG~\cite{fakecatcher} & 94.87 & 95.75 & 91.07\\
    \midrule
    
     Inception~\cite{inception} & 65.5* & 54.4*   & 68.88\\

    Xception~\cite{xceptionnet} & 74.5*  & 70.9*  & 75.55\\
\midrule
     Ours & 93.28 & 91.62 & 80.00\\
   \bottomrule
 \end{tabular}
    \end{table}

\begin{figure}[h]
    \centering
        \includegraphics[width=1\linewidth]{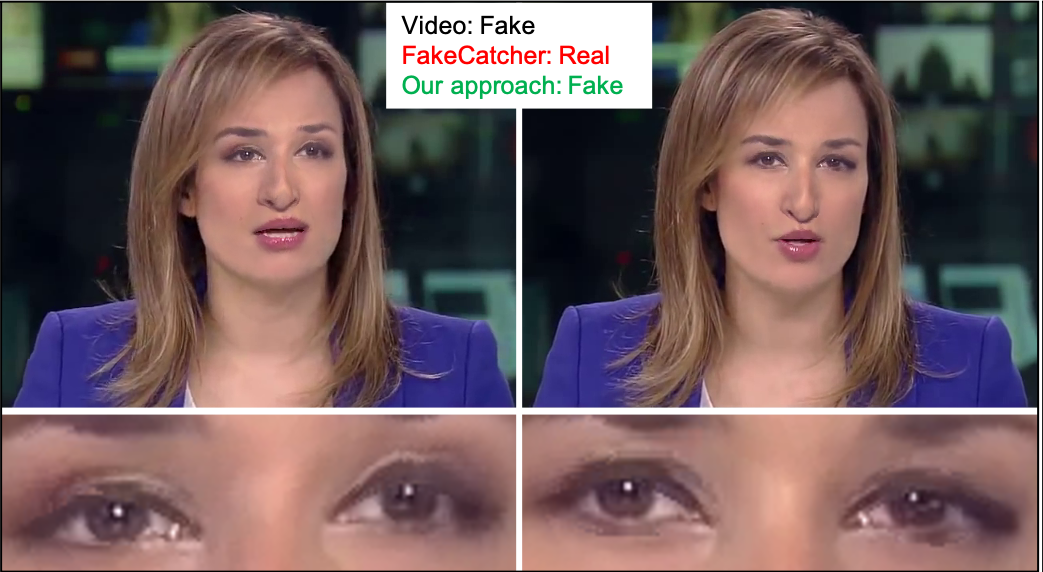}

\caption{\textbf{Comparison to FakeCatcher~\cite{fakecatcher}.} A fake video classified as real by~\cite{fakecatcher} and as fake by our approach. Artifacts are more visible when zoomed in (bottom). Fake frames from~\cite{FF++}. \\} 
  \label{fig:fc}
\end{figure}

Based on the slightly better results, we investigate and compare~\cite{fakecatcher} and our approach. We observe that the failure cases are mutually exclusive, as demonstrated in Figure~\ref{fig:fc}. For false negatives (left half), PPG signals extracted from the skin may be smooth enough to trick~\cite{fakecatcher} to classify a fake video as real, however, our approach correctly classifies the same fake video with 84.93\% confidence from the eye artifacts (bottom row). For false positives (right half), heavy make up on the skin may deteriorate PPG signals and lead~\cite{fakecatcher} to detect a real as fake, however, our approach correctly classifies it due to coherent eye signatures. This motivates us to propose ensemble networks, merging biological detectors working in separate domains to cover for each others' limitations.

\subsubsection{Comparison with Other Networks}
We also compare our approach against six standard deep network architectures, without integrating our eye and gaze features. In Table~\ref{tab:sota}, we listed~\cite{inception} and~\cite{xceptionnet} results on three datasets, and in Table~\ref{tab:compbio}, we listed~\cite{mesonet} and~\cite{shallownet} variation results in addition to those of two basic architectures, a three-layer CNN and a ConvLSTM. Our motivation in these comparisons is to justify that the eye and gaze features we formulate contain sufficient information to enable our simple architecture to perform better than deeper and more complex models with more capacity and full frame information. 
\begin{table}[h]
\centering
\caption{\textbf{Comparison} with other deep fake detectors on DF and FF datasets. Same train/test data and procedure are used for all, except marked results~(*) from~\cite{FF++}.}
  \label{tab:sota}
  \begin{tabular}{ccc} 
    \toprule
    Approach & Dataset & Acc.\\
    \midrule
    3-layer CNN & DF& 48.88 \\
    ConvLSTM & DF& 48.83\\
    \cite{shallownet} V3 & DF& 73.33\\
    \cite{shallownet} ensemble & DF& 80.00\\
    MesoNet~\cite{mesonet} & FF/DF &87.3* \\ 
    MesoNet~\cite{mesonet} & FF/FS &61.2* \\ 
    \bottomrule
\end{tabular}

\end{table}

We obtained the results on FF++ from their benchmark, concluding with our approach having 5.98\% better accuracy than the second best~\cite{mesonet}, verifying the contribution of our authenticity formulation. For the results on DF, we use the same setup for training all of the models, namely, the same train/test splits with the same hyper parameters listed in Section~\ref{sec:training}. The only difference is that others are trained with source frames (real or fake) without preprocessing, and ours is trained with our signatures. Despite the in-the-wild setting of DF dataset, our approach outperforms all listed detectors -- except having the same accuracy with the complex ensemble network of~\cite{shallownet} utilizing full frame information with two order of magnitude more parameters than our model. We emphasize that we \textit{only} input eyes in our fake detector, and still obtain better results, proving the power of the proposed eye and gaze features.

\subsubsection{Cross Validation}
\label{sec:fold}
Giving the benefit of doubt to our random train/test split of datasets, we perform two 5-fold validations with different splits. In the first one, we create five different train-test subset pairs by randomly splitting the aforementioned FF++ dataset as 70\%-30\%. Then we train and test our approach independently five times on these subsets, with the same procedure discussed in Section~\ref{sec:training}. Five test accuracies have a mean of 90.6\% with 1.5\% standard deviation.

In the second validation test, we use 80\%-20\% split to create non-overlapping subsets. In other words, 2000 training videos (1000 fake, 1000 real) are randomly split to mutually exclusive subsets of 400 videos (200 fake, 200 real) as test sets. For each test set, remaining 1600 videos are used for training. Five train and test sessions yield 90.3\% mean accuracy with 1.1\% standard deviation, validating that our results are independent of the random dataset splits.

Lastly, we conduct a cross dataset evaluation between CDF and DFor datasets. Using the same train and test setup, we train a model on CDF and test on DFor, then we train a model on DFor and test on CDF. We obtain 80.72\% and 73.68\% accuracies respectively, which are close to our in-the-wild results.

\subsection{Experiments}
\label{sec:Experiments}
We summarize the effects of different sequence durations, using several feature sets, under two image-space artifacts.

\subsubsection{Sequence Duration}
As our method benefits from both temporal and spectral artifacts of continuous signals, such as 3D gaze points in time or frequency bands of iris colors, selection of $\omega$ is crucial. We experimentally validate our selection of $\omega$ with an ablation study (Table~\ref{tab:Omega}), testing different sequence lengths of 16, 32, 64, and 128 on FaceSwap subset of FF++. We monitor that short sequences may contain only noise or missing faces, while long sequences saturate authenticity cues. We can recover from the former during video prediction, but not from the latter one. In short, we conclude on the optimum value of $\omega=32$ from Table~\ref{tab:Omega}.
\begin{table}[h]
\centering
   \caption{\textbf{Sequence Durations.} We validate the optimum choice of $\omega=32$ by sequence and video accuracies.}
  \label{tab:Omega}
  \begin{tabular}{c|cccc}
    \toprule
    $\omega$ & 16& 32& 64& 128\\
    \midrule
    S. Acc. & 77.72 & 80.63& 79.79 & 78.60\\
    V. Acc. &89.27&91.62&86.93&85.04\\
  \bottomrule
\end{tabular}
\end{table}

\subsubsection{Feature Sets}
Next, we train and test our system on FF++ dataset with different feature collections, namely (i) as is, only with (ii) spectral, (iii) temporal, (iv) geometric, (v) visual, and (vi)  metric features, (vii-ix) only with raw gaze variants, (x) without metric features, and (xi) without geometric features. All other hyper parameters are held the same. According to Table~\ref{tab:Ablation}, all five domains of features are contributing to the detection accuracy, however metric features have relatively less importance. Although metric-only results are high (row 6), results without metric features are higher (row 10). We speculate that the size of metric features are relatively smaller in the signature, so they influence the classification less than others, and they can be compensated by other features.

\begin{table}[h]
        \centering
  \caption{\textbf{Ablation study} with and without some of the feature domains from Section~\ref{sec:features}.}
  \label{tab:Ablation}
  \begin{tabular}{ccc}
    \toprule
    Condition& S. Acc. & V. Acc.\\
    \midrule
    All         &83.12      & 92.48\\
    Spec-only   &75.42      & 84.46\\
    Temp-only   &76.39      & 85.28\\
    Geo-only    &79.84      & 91.82\\
    Visual-only &78.56      & 88.46\\
    Metric-only &78.52      & 87.79\\
    Raw gaze   & 72.88  & 81.04 \\
    Gaze vectors & 72.28 & 80.80\\ 
    Rows 6 and 7 & 72.66 & 81.39\\
    No metric   &79.45      &88.96\\
    No geometric&79.52      &87.98\\ 
  \bottomrule
\end{tabular}
\end{table}

\subsubsection{Postprocessing Artifacts}
Generative noise is not the only artifact that deep fakes exhibit, especially in the wild. Post-processing operations, such as compression and resampling, may deteriorate focused signals. In order to test the robustness of our approach, we approximate the post-processing artifacts with two commonly used image processing operations: Gaussian blur and median filtering. All other parameters are kept the same. As enumerated in Table~\ref{tab:Blur}, our system is robust against Gaussian and median noise, because effects of such artifacts are minimal for the eye region. As long as the eyes are still visible in videos, features from different domains enforces the detection accuracy to remain the same -- i.e., high resolution eyes enable the contributions of geometric features, and low resolution eyes enable the contributions of visual features. This experiment supports the claim that eye/gaze based fake detection is more persistent when other biological fake detectors fail under postprocessing operations, compared to Table 14 of~\cite{fakecatcher}. Their accuracy drops by 20.66\% under 9x9 Gaussian blur, whereas our accuracy only drops by 0.37\%.
\begin{table}[h]
\centering
  \caption{\textbf{Robustness.} We manually add Gaussian blur and median noise to input videos and evaluate our robustness against such post-processing artifacts.}
  \label{tab:Blur}
  \begin{tabular}{c|c|ccc|ccc}
  \toprule
 & Initial & \multicolumn{3}{c|}{Gaussian Blur} & \multicolumn{3}{c}{Median Filter} \\
\midrule
Kernel    & N/A      & 3x3       & 7x7       & 9x9       & 3x3       & 7x7       & 9x9       \\
Accuracy  & 92.48    & 89.13     & 85.88     & 92.11     & 91.27     & 90.77     & 89.76\\
\bottomrule
\end{tabular}
\end{table}

\section{Conclusion and Future Work}
In this paper, we exhaustively analyze real and fake gazes to answer \textit{"Where do Deep Fakes Look?"}, and morph our findings into a deep fake detector. To the best of our knowledge, our paper conducts the first extensive analysis of deep fake gazes, and it is the first approach to build a detector solely based on holistic eye and gaze features (instead of cherry-picking a few). We evaluate our approach on four datasets, compare against biological and deep detectors, and conduct several ablation studies. As eyes in real videos exhibit natural signatures, which fake ones yet to mimic consistently, we put forward that our visual, geometric, temporal, metric, and spectral features to be integrated to existing fake detectors.

One advantage of our approach is that our signatures are based on authenticity signals instead of generative noise. Blind classifiers without "real" priors are prone to adversarial attacks, as shown in \cite{Hussain_2021_WACV} for \cite{xceptionnet} and~\cite{mesonet}. In contrast, fake eye/gaze physics should be spatio-temporally accurate in order to fool our classifier.

On the other hand, as the arms race continues, we also believe that our findings can be used to create better deep fakes with consistent fake gazes. Such synthetic gazes with increased photorealism can be useful for virtual/augmented reality applications, avatars, data augmentation, transfer learning, and controlled learning and testing environments for eye tracking research. 

Another future direction is to include gaze classification for our 3D gaze points in order to boost the accuracy of our fake detector. We observe that time-varying 3D gaze points contain much richer biological information that can be utilized if we can interpret them further -- starting from gaze movements such as saccades, fixations, etc. We anticipate that our first step in deciphering fake gazes will lead the way for other deep learning approaches to be inherited for eye or gaze based deep fake detection techniques.

%\begin{acks}
%To Robert, for the bagels and explaining CMYK and color spaces.
%\end{acks}

\bibliographystyle{ACM-Reference-Format}
\bibliography{References}

\appendix
\section{Signature Composition}\label{app:a}
In Table~\ref{tab:sign}, we list elements of our input tensor (i.e., the signature) that collates the spectral and temporal versions of the aforementioned visual, geometric, and metric features. SS means ``shift by min+scale by max'' and $d^+$ corresponds to maximum IPD. Multiple features occupying one row corresponds to features being merged into a three channel value. Channels with stars (*) indicate that a single channel value is duplicated into multiple channels per its importance. Cross correlation features intake double signals and output single signals. The signatures include 24 visual, 24* geometric, and 12 metric features. Finally, all $|T|=20 \times 3 \times \omega$ values are transformed and resized to the spectral domain, creating $|S|=20 \times 3 \times \omega$ values, constituting the $40\times \omega \times 3$ size signatures.

\begin{table}[hb]
\centering
  \caption{\textbf{Signatures.} We enlist the elements, their parts, sides, channels, and normalizations that make up our fake and real signatures.}
  \label{tab:sign}
  \begin{tabular}{cccccc}
    \toprule
    Set&Feature&Part&Side&Chan.&Norm.\\
    \midrule
    Visual & Color & Iris & Left & 3 & /256\\\hline
    Visual & Color & Iris & Right & 3 & /256\\\hline
    Visual & Color & Pupil & Left & 3 & /256\\\hline
    Visual & Color & Pupil & Right & 3 & /256\\\hline
    Visual & Area & Eye & Left & 1 & SS\\
    Visual & Area & Eye & Right & 1 & SS\\
    Geometric & Area & Eye & Left-Right & 1 & $/d^+$\\\hline 
    Visual & Area & Iris & Left & 1 & SS\\
    Visual & Area & Iris & Right & 1 & SS\\
    Geometric & Area & Iris & Left-Right & 1 & SS\\\hline
    Visual & Area & Pupil & Left & 1 & SS\\
    Visual & Area & Pupil & Right & 1 & SS\\
    Geometric & Area & Pupil & Left-Right & 1 & $/d^+ -1$\\\hline 
    Visual & Color & Iris & Left-Right & 3 & /256\\\hline
    Visual & Color & Pupil & Left-Right & 3 & /256\\\hline
    Geometric & Vector & Gaze & Left & 3 & SS2\\\hline
    Geometric & Vector & Gaze & Right & 3 & SS2\\\hline
    Geometric & Point & Gaze & - & 3 & SS\\\hline 
    Geometric & Error & Gaze & - & 3 & $/d^+$\\\hline
    Geometric & Cost & Gaze & - & 3* & SS\\\hline
    Geometric & Distance & Eye & - & 3* & SS\\\hline
    Geometric & Distance & Pupil & - & 3* & SS\\\hline
    Cross & Color & Iris & Both & 2x3 > 3 & SS\\\hline
    Cross & Color & Pupil & Both & 2x3 > 3 & SS\\\hline
    Cross & Area & Iris & Both & 2 > 1 & SS\\\
    Cross & Area & Pupil & Both & 2 > 1 & SS\\\
    Cross & Area & Eye & Both & 2 > 1 & SS\\\hline
    Cross & Vector & Gaze & Both & 2x3 > 3 & SS\\
  \bottomrule
\end{tabular}

\end{table}
\section{Network Layers}\label{app:b}
We log our network layers, their arguments, and their output shapes in Table~\ref{tab:layers}. Note that the order of layers and the network architecture are finalized empirically and theoretically~\cite{disharmony,deeppyr}. We would also like to note that the main motivation of our paper is to analyze eye and gaze behavior of various deep fakes, and this simple but strong network is a part of the verification process for our findings. For production-level accuracies, we encourage researchers to invest in neural architecture search for building better alternatives utilizing our eye and gaze features.

\begin{table}[h]
  \caption{\textbf{Network Layers.} We document sizes, arguments, and outputs of each layer in our network.}
  \label{tab:layers}

  \begin{tabular}{ccc}

\hline
Layer     & Arguments  & Output\\
\hline
Signature           &-      &40x64x3\\
\hline
Flatten           &-      &7680x1x1\\\hline
Batch Norm      &-     &7680\\
Dense & 256  & 256\\\hline
Batch Norm      &-     &256\\
Leaky ReLu      &0.2     &256\\
Dropout      &0.3     &256\\
Dense & 128  &128\\\hline
Batch Norm      &-     &128\\
Leaky ReLu      &0.2     &128\\
Dropout     &0.3     &128\\
Dense & 64  & 64\\\hline
Dense & 2  & 2 \\
Sigmoid & -  & 2\\

\hline
\label{tab:network}
\end{tabular}

\end{table}

\end{document}